\documentclass[sigconf]{acmart}

\usepackage{booktabs}
\usepackage{multirow}
\usepackage{algpseudocode}
\usepackage{caption}
\usepackage[T1]{fontenc}
\usepackage{array} 
\usepackage{algorithm}
\usepackage{graphicx}
\usepackage{tabularx}
\usepackage{makecell}
\usepackage{xcolor}
\usepackage{bm}
\usepackage{amsmath} 
\usepackage{alphalph}

\AtBeginDocument{%
  }

\setcopyright{acmlicensed}
\copyrightyear{2026}
\acmYear{2026}
\acmDOI{XXXXXXX.XXXXXXX}
\renewcommand\footnotetextcopyrightpermission[1]{} 
\acmConference[MM '26]{Proceedings of the 34th ACM International Conference on Multimedia}{November 10-14, 2026}{Rio de Janeiro, Brazil}
\acmISBN{978-1-4503-XXXX-X/2018/06}

\acmSubmissionID{4882}

\begin{document}

\title{\textbf{Long-CODE}: Isolating Pure \textbf{Long}-\textbf{C}ontext as an \textbf{O}rthogonal \textbf{D}imension in Video \textbf{E}valuation}

\author{Zhijiang Tang}
\authornote{Both authors contributed equally to this research.}
\email{tangzhijiang24@mails.ucas.ac.cn}
\affiliation{%
  \institution{Hangzhou Institute for Advanced Study, UCAS}
  \country{China}
}

\author{Jiaxin Qi}
\authornotemark[1]
\email{jxqi@cnic.cn}
\affiliation{%
  \institution{Computer Network Information Center, CAS}
  \country{China}
}

\author{Bing Zhao}
\email{zhaobingchs@gmail.com}
\affiliation{%
  \institution{Department of AI Infrastructure, Bilibili Inc.}
  \country{China}
  }

\author{Jianqiang Huang}
\authornote{Corresponding author.}
\email{jqhuang@cnic.cn}
\affiliation{%
  \institution{Computer Network Information Center, CAS}
  \country{China}
}
\affiliation{%
  \institution{Hangzhou Institute for Advanced Study, UCAS}
  \country{China}
}

\renewcommand{\shortauthors}{Trovato et al.}

\ccsdesc[500]{Computing methodologies~Computer vision}
\keywords{Long Video Evaluation, Human Consistency}


\received{20 February 2007}
\received[revised]{12 March 2009}
\received[accepted]{5 June 2009}

\begin{abstract}
As video generation models achieve unprecedented capabilities, the demand for robust video evaluation metrics becomes increasingly critical. Traditional metrics are intrinsically tailored for short-video evaluation, predominantly assessing frame-level visual quality and localized temporal smoothness. However, as state-of-the-art video generation models scale to generate longer videos, these metrics fail to capture essential long-range characteristics, such as narrative richness and global causal consistency. Recognizing that short-term visual perception and long-context attributes are fundamentally orthogonal dimensions, we argue that long-video metrics should be disentangled from short-video assessments.
In this paper, we focus on the rigorous justification and design of a dedicated framework for long-video evaluation. We first introduce a suite of long-video attribute corruption tests, exposing the critical limitations of existing short-video metrics from their insensitivity to structural inconsistencies, such as shot-level perturbations and narrative shuffling. To bridge this gap, we design a novel long-video metric based on shot dynamics, which is highly sensitive to the long-range testing framework. Furthermore, we introduce Long-CODE (Long-\textbf{C}ontext as an \textbf{O}rthogonal
\textbf{D}imension for video \textbf{E}valuation), a specialized dataset designed to benchmark long-video evaluation, with human annotations isolated specifically to genuine long-range characteristics. Extensive experiments show that our proposed metrics achieve state-of-the-art correlation with human judgments. Ultimately, our metric and benchmark seamlessly complement existing short-video standards, establishing a holistic and unbiased evaluation paradigm for video generation models.
\end{abstract}
\maketitle

\section{Introduction}

\begin{figure}
    \centering
    \includegraphics[width=1\linewidth]{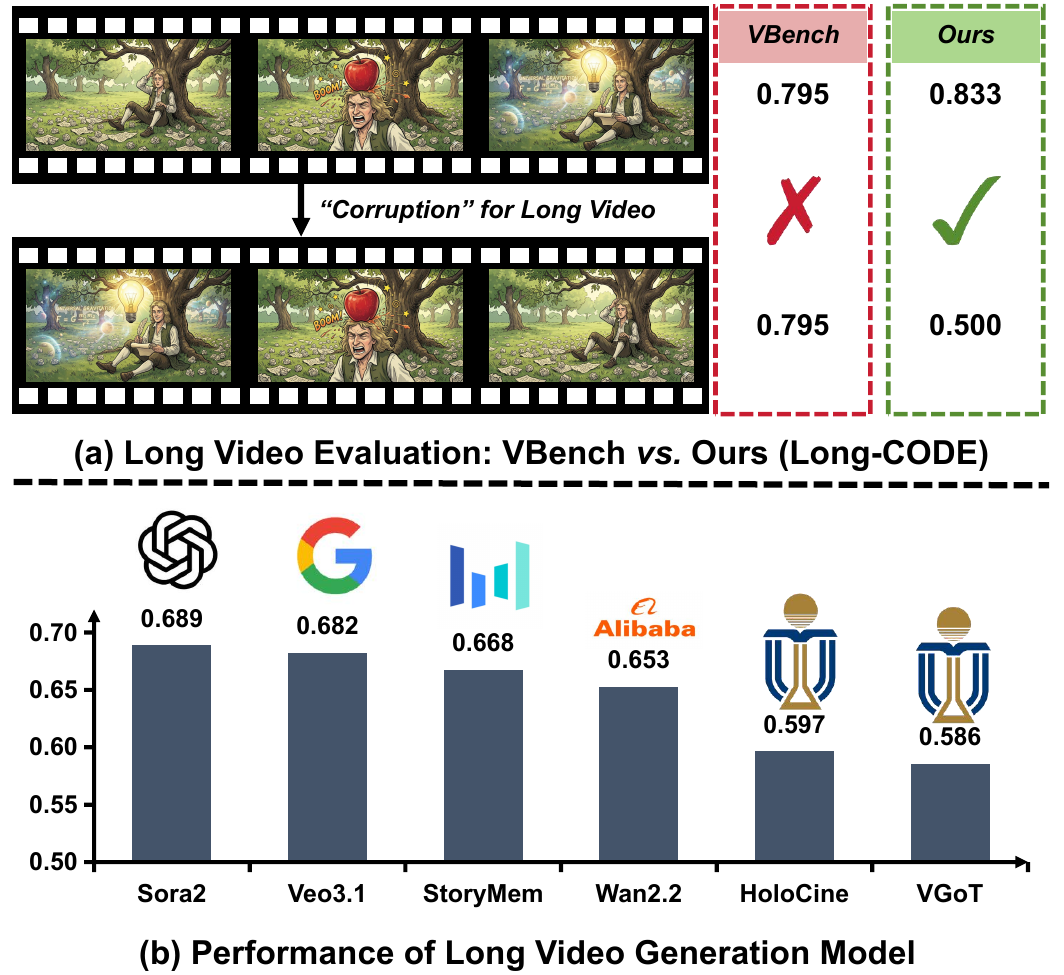}
    \caption{
Illustrations of our proposed Long-CODE benchmark. 
    (a) Comparison between the short video metric VBench~\cite{vbench} and our proposed metric, Long-CODE. The "Corruption" test involves shuffling the original shot order of a long video depicting the folklore of Newton discovering gravity~\cite{brewster2010memoirs}.
    (b) Performance benchmarking of cutting-edge long video generation models, including OpenAI's Sora 2~\cite{sora}, Google's Veo 3.1~\cite{veo}, ByteDance's StoryMem~\cite{storymem}, Alibaba's Wan 2.2~\cite{wan}, and HKUST's HoloCine~\cite{holocine} and VGoT~\cite{vgot}. Please refer to Section ~\ref{sec:exp} for the generation pipeline.
}
    \label{fig:teaser}
\end{figure}

Recent advancements in Video Generation Models (VGMs) have yielded unprecedented visual fidelity and realism~\cite{sora,veo, wan}. To systematically evaluate the synthesis capabilities of these models, various metrics and benchmarks have been established~\cite{li2025worldmodelbench}, such as VBench~\cite{vbench} for frame-level visual quality, VideoCLIP~\cite{videoclip} for text-to-video semantic alignment, and LocoT2V-Bench~\cite{zheng2025locot2v} for localized temporal smoothness.

However, existing metrics are intrinsically tailored for short-video evaluation, typically spanning only a few seconds~\cite{vbench,liu2024evalcrafter}. As state-of-the-art VGMs scale to generate minute-level content~\cite{storymem,vgot}, evaluating video quality necessitates dimensions far beyond short-term perception. Specifically, long-video generation introduces additional higher-order attributes, such as narrative richness and long-term causal consistency. For example, whether an object that exists in the camera view logically reappears later with consistent physical properties. These long-range cognitive and structural evaluations fundamentally exceed the functional scope of existing short-video metrics, which are inherently limited to capturing adjacent frame variances or localized semantic distances. 
As illustrated in Figure~\ref{fig:teaser}(a), applying temporal corruption to a long video may preserve the visual quality and local smoothness of individual shots. Consequently, traditional short-video metrics, such as VBench, fail to register a decline in performance. However, such a video is fundamentally flawed for long-form content because its underlying narrative structure and causal logic have been destroyed.

To bridge this gap, we argue that long-video attributes and short-video visual perception are fundamentally orthogonal dimensions. Traditional metrics remain valid and necessary for guaranteeing the foundational visual quality of long videos, which can be easily achieved through sliding-window or aggregation strategies. Therefore, what the community urgently requires is a dedicated evaluation paradigm that is maximally decoupled from baseline visual quality, focusing exclusively on the emergent characteristics of long-context generation. To this end, we introduce \textbf{Long-CODE} (\textbf{Long}-\textbf{C}ontext as an \textbf{O}rthogonal \textbf{D}imension in Video \textbf{E}valuation), a novel framework designed to seamlessly complement existing standards and establish a holistic evaluation ecosystem.


In this paper, we present a systematic approach to long-video evaluation. 
First, we formally define a long-context video as an organic sequence of multiple interconnected shots, where a traditional short video is merely a single constituent element of this set. Furthermore, we provide a theoretical justification demonstrating that long-video attributes are orthogonal to short-video attributes, necessitating a distinct evaluative dimension. 

Based on this formulation, we design a suite of long-video attribute-corruption tests (e.g., shot shuffling and replacement) and empirically demonstrate the critical insensitivity of existing short-video metrics to these long-range structural perturbations, thereby justifying their limitations for long-range video evaluation. 
Second, we propose \textbf{Dynamic Structure Alignment}, a novel evaluation metric specifically designed to capture long-range dependencies and temporal dynamics across multiple shots.
Within this framework, we define {structural similarity vectors} to represent the inter-shot relationships. We utilize the Spearman rank correlation coefficient~\cite{spearman} to evaluate the alignment between the generated video vector and the corresponding prompt vector. To complement the assessment of long-video semantics, we integrate Multimodal Large Language Models (MLLMs) to perform ``deep thinking'' and provide qualitative scoring. The final evaluation score is derived from the fusion of these two distinct sub-metrics.

Finally, we construct the \textbf{Long-CODE} benchmark, a specialized dataset comprising complex, multi-shot prompts. Specifically, we leverage an MLLM to decompose a complex storyline into a sequence of structured shots, thereby yielding the individual shot prompts. We also introduce a rigorous human evaluation protocol strictly isolated to gauge genuine long-range characteristics. 
For instance, during the annotation process, human evaluators are instructed to prioritize narrative fluidity over image quality, ensuring the benchmark only reflects true long-context performance.
Extensive experiments validate the effectiveness of our proposed framework. Our metric exhibits high sensitivity in the designed corruption tests, yielding orthogonal results to standard short-video metrics, validating its unique utility. Furthermore, evaluated on the Long-CODE benchmark, our proposed metric achieves state-of-the-art correlation with human judgments specifically targeted at long-video quality. Ultimately, our metric and benchmark serve as a robust complement to existing short-video standards, offering a comprehensive and unbiased evaluation paradigm for current video generation models.

We summarize our primary contributions as follows:
\begin{enumerate}
    \item We critically analyze the limitations of current video evaluation paradigms, demonstrating their systematic failure to capture long-range video. By establishing the {orthogonality} between short-term visual perception and long-term contextual attributes, we advocate for a decoupled evaluation framework and propose the first dedicated methodology focused exclusively on the structural integrity of long videos.

    \item We introduce an {attribute corruption framework} that exposes the insensitivity of prevailing metrics to macro-level logical inconsistencies. To address this, we develop two novel metrics based on shot dynamics and narrative flow. Empirical results reveal a performance divergence, where our metrics capture errors that short-video benchmarks overlook, thereby validating our hypothesis regarding the dimensional independence of long-form video quality.

    \item We curate and release {Long-CODE}, a specialized benchmark dataset designed to isolate and evaluate genuine long-range characteristics. Through extensive human experiments, we demonstrate that our proposed metrics achieve state-of-the-art alignment with human evaluation, providing a necessary and robust complement to existing standards for a truly holistic assessment of video generation models.
\end{enumerate}
\textbf{Code: }\href{https://github.com/ZhijiangTang/Long-CODE}{https://github.com/ZhijiangTang/Long-CODE}  

\section{Related Work}
\subsection{Video Generation Models}
With the rapid development of diffusion models~\cite{ho2020denoising, lipman2022flow, peebles2023scalable}, Video Generation Models (VGMs) have achieved astonishing visual results. For short-video generation, the open-source community has made significant strides~\cite{zheng2024open,huang2024owl}: VDM~\cite{ho2022video} extends diffusion architectures to the video domain; CogVideoX~\cite{yang2024cogvideox} leverages 3D causal VAEs and expert transformers for efficient synthesis; the Wan series~\cite{wan} provides highly competitive, open-source, high-fidelity generation capabilities. Meanwhile, closed-source models such as Sora~\cite{sora} and Veo~\cite{veo} have set new industry benchmarks by producing remarkably realistic, physically grounded short clips.
Beyond short clips, there is a growing trend towards the generation of long videos~\cite{vgot,zhou2024storydiffusion}. Frameworks like StoryMem~\cite{storymem} incorporate memory mechanisms to preserve character and scene consistency over extended durations. Similarly, HoloCine~\cite{holocine} proposes a holistic generation paradigm specifically designed to ensure cinematic multi-shot narrative coherence. In addition to these standalone systems, various plug-and-play modules~\cite{cai2025mixture,jia2025moga} have been developed to seamlessly empower existing short-video models with long-horizon storytelling capabilities.

\subsection{Video Benchmarks}
Alongside the evolution of VGMs, numerous video benchmarks have been developed to quantify their capabilities. VBench~\cite{vbench,huang2025vbench++} provides a comprehensive, fine-grained, multidimensional metric suite for generic visual quality. To further assess adherence to physical laws, enhanced benchmarks such as VBench-2.0~\cite{zheng2025vbench} and WorldModelBench~\cite{li2025worldmodelbench} have been introduced to evaluate how well models simulate real-world physical dynamics. Another emerging paradigm utilizes Multimodal Large Language Models (MLLMs) for automated scoring; for instance, VideoQA~\cite{song2026vqqa,lin2024evaluating}, Video-Bench~\cite{videobench}, and VideoScore2~\cite{videoscore2} employ advanced vision-language models to semantically evaluate generated content.
Recently, new research has begun to explore long-video benchmarks, such as LocoT2V~\cite{zheng2025locot2v}. However, the multidimensional metrics proposed in these early attempts are essentially extrapolations of short-video metrics and fundamentally fail to genuinely assess the intrinsic narrative and structural quality of long videos. In this paper, we theoretically and experimentally demonstrate that short-video metrics and long-video metrics operate on orthogonal dimensions when evaluating long-form generation. Therefore, we propose Long-CODE as a dedicated and necessary complement to existing short-video standards, with extensive experiments strongly validating the effectiveness and indispensability of our benchmark.

\begin{figure*}[htbp]
    \centering
    \includegraphics[width=1\linewidth]{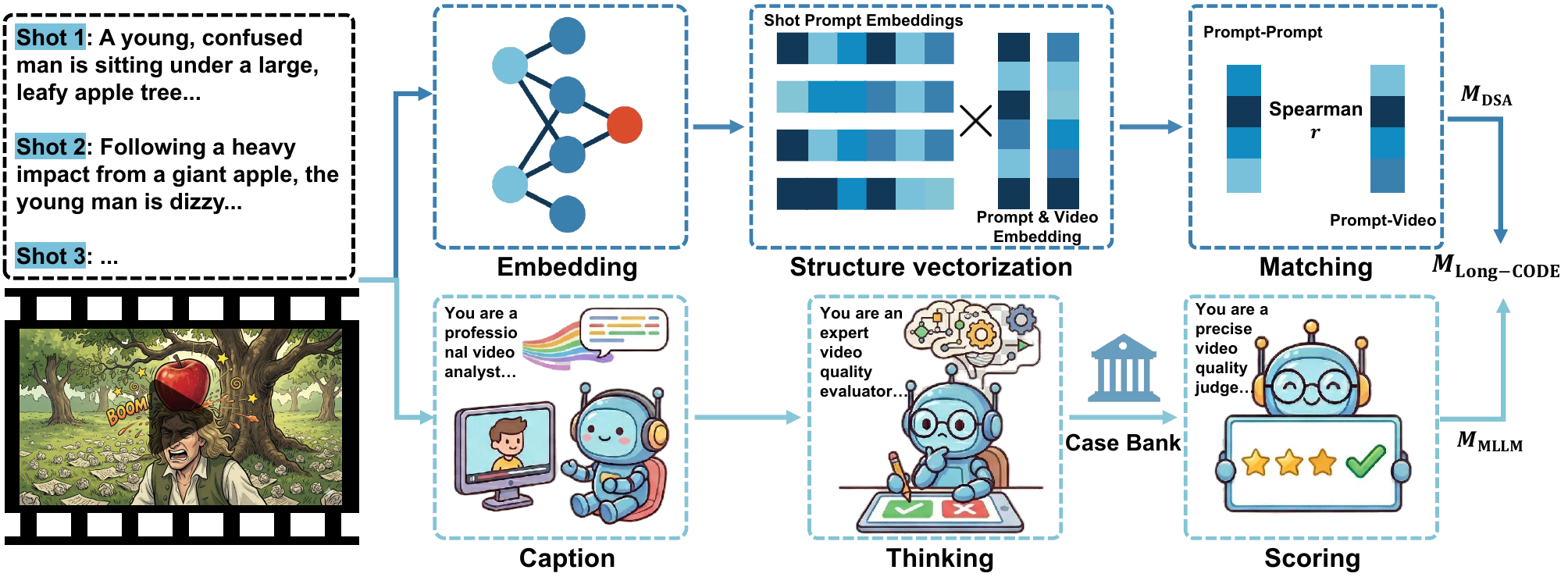}
    \caption{
{The framework of Long-CODE.} The shot prompts and the generated video together are processed through two parallel and complementary evaluation pipelines. The upper branch calculates the structural dynamic score ($M_\text{DSA}$) by 
extracting embeddings from both modalities, performing structure vectorization, and 
computing the matching correlation (i.e., via Spearman's $r$) between prompt-prompt and prompt-video structural similarity vectors to assess structural consistency. Concurrently, the lower branch utilizes an MLLM to compute the semantic quality score ($M_{MLLM}$) through a structured, multi-stage workflow encompassing ``Captioning'', ``Thinking'', and ``Scoring''. Finally, $M_\text{DSA}$ and $M_\text{MLLM}$ are aggregated to formulate the comprehensive long-video evaluation metric, $M_{\text{Long-CODE}}$
    }
    \label{fig:method}
\end{figure*}
\section{Method}


\subsection{Preliminary}
\noindent\textbf{Conditional Flow Matching~(CFM)}~\cite{lipman2022flow}. Modern video generation models commonly employ CFM as the training objective~\cite{zheng2024open,yang2024cogvideox}. Given a noise sample $x_0 \sim \mathcal{N}(0, I)$ and a target sample $x_1\sim\mathcal{D}$ from dataset $\mathcal{D}$, the interpolant is defined as $x_t = (1-t)x_0 + tx_1$ for $t \in [0,1]$. A neural network $v_\theta$ is trained to predict the velocity field:
\begin{equation}
    \mathcal{L}_{\text{CFM}}(\theta) = \mathbb{E}_{t,\, x_0\sim \mathcal{N},\, x_1\sim\mathcal{D}}\left[\left\| v_\theta(x_t, t) - (x_1 - x_0) \right\|^2\right] 
\end{equation}
At inference, the generated sample is obtained by integrating the learned ODE $\mathrm{d}x_t/\mathrm{d}t = v_\theta(x_t, t)$ from $t{=}0$ to $t{=}1$~\cite{lipman2022flow,liu2025flow}.

\noindent\textbf{Shot-Based Long Video Generation}. State-of-the-art long video generation models~\cite{storymem,vgot} decompose the generation task into a sequence of shot-level productions. Formally, a long video $\mathcal{V}$ is produced as the temporal concatenation of $K$ shots:
\begin{equation}
  \mathcal{V} = s_1 \oplus s_2 \oplus \cdots \oplus s_K,\quad s_k = G_\theta\bigl(c_k,\;\mathcal{S}_{<k}\bigr)  
\end{equation}
where $c_k$ denotes the text prompt for the $k$-th shot, $\mathcal{S}_{<k} = \{s_1, \ldots, s_{k-1}\}$ represents all previously generated shots that provide conditioning context (e.g., the last frame of $s_{k-1}$ serves as the initial frame for generating $s_k$), and $\oplus$ denotes temporal concatenation. Each shot $s_k$ is itself a short video clip generated via a flow-matching based diffusion model conditioned on both the textual description and visual context from preceding shots.

For models lacking native long-video generation capabilities (e.g., Sora and Veo), a prevalent paradigm involves a shot-by-shot synthesis approach. Specifically, each shot is generated by conditioning on the final frame of the preceding shot and the text prompt for the current shot. Ultimately, these discrete shots are temporally concatenated to synthesize a cohesive long video.

\subsection{Long-Context Video Attributes}
\noindent\textbf{Theoretical Analysis.} We define a long-context video $\mathcal{V}$ as an ordered sequence of shots $s_1, s_2, \ldots, s_K$, where each shot $s_k$ is a short video clip corresponding to a specific scene or narrative segment. Unlike short videos, which can be fully characterized by frame-level visual quality and local temporal smoothness, long videos inherently carry additional long-range structural attributes that emerge exclusively from inter-shot relationships~(e.g., temporal logic and causal consistency).

We denote by $M_S$ a {short-video metric} that evaluates individual shot quality. To score a long video $\mathcal{V}$, existing short-video benchmarks~(e.g., VBench~\cite{vbench}) typically aggregate per-shot scores through a symmetric aggregator $\phi$~(e.g., arithmetic mean):
\begin{equation}\label{eq:ms}
  M_S(\mathcal{V}) = \phi\bigl(q(s_1),\, q(s_2),\, \ldots,\, q(s_K)\bigr),
\end{equation}
where $q\colon \mathcal{S} \to \mathbb{R}$ maps each shot to a scalar quality score. 

A {long-context metric} $M_L$, by contrast, is structure-aware: it is sensitive to the sequential inter-shot coherence encoded by the pairwise relation matrix $\mathbf{R} \in \mathbb{R}^{K \times K}$, where $R_{ij} = r(s_i, s_j)$ measures the semantic or temporal coherence between shots $i$ and $j$.

\begin{proposition}\label{prop:orth}
For the short-video metric $M_S$ defined in Eq.~\eqref{eq:ms} with a symmetric aggregator $\phi$, and a long-context metric $M_L$ that depends solely on the inter-shot relation matrix $\mathbf{R}$, the mutual information between $M_S$ and $M_L$ vanishes:
$$I(M_S ; M_L) = \iint p(M_S, M_L) \log \frac{p(M_S, M_L)}{p(M_S)\,p(M_L)}\, dM_S\, dM_L = 0.$$
\end{proposition}

The proof of this proposition is in the supplementary material. This result establishes that any aggregation-based short-video metric is provably insensitive to structural perturbations in the shot sequence, underscoring the need to introduce a complementary long-context evaluation dimension. We empirically validate this orthogonality using the following corruption-testing framework.

\noindent\textbf{Corruption Testing Framework.} We design four families of corruption operators that target long-range structural attributes (influencing $M_L$) while reducing per-shot visual quality degradation (preserving $M_S$), thereby exposing the insensitivity of existing short-video metrics:
\begin{enumerate}
    \item \textbf{Shuffle}. The shot ordering of $\mathcal{V}$ is randomly permuted according to a uniformly sampled permutation $\pi$. Each shot's visual content remains completely intact.

    \item \textbf{Replace}. A subset of constituent shots within a video $\mathcal{V}$ is replaced by exogenous shots retrieved from a comprehensive {shots Bank} (constructed from large-scale open-source video repositories). 

    \item \textbf{Edition}. A subset of shots in $\mathcal{V}$ is re-edited by a depth-guided video editing model~(i.e., VACE~\cite{jiang2025vace}), which replaces key semantic elements~(e.g., character gender or identity) while preserving the spatial layout and motion dynamics. 

    \item \textbf{Synthesis}. A subset of shots in $\mathcal{V}$ is regenerated from modified descriptions: an MLLM first captions the original shot, the caption is then rewritten to be surreal, and VGM synthesizes a replacement shot from the altered prompt and the original first frame. 
\end{enumerate}
All corruption operators are parameterized by a strength hyperparameter that controls the intensity of the perturbation~(e.g., the number or fraction of shots affected). 
Extensive experiments in Section~\ref{sec:exp} confirm that representative short-video metrics remain largely invariant under these corruptions, empirically validating the orthogonality established in Proposition~\ref{prop:orth}.

\subsection{Long-CODE}

\noindent\textbf{Dynamic Structure Alignment (DSA)}. 
DSA measures the structural alignment between intended and realized shot dynamics through a video-language embedding space. Given a video $\mathcal{V}$ with shot prompts $\mathcal{C} = \{c_1, \ldots, c_K\}$, let ${f}_t(\cdot)$ and ${f}_v(\cdot)$ denote the L2-normalized text and video embedder, respectively. We define two $K$-dimensional similarity vectors:
\begin{equation}
    s_i^{c} = {f}_t(c_i)^\top\, {f}_t(\mathcal{C}),\quad s_i^{v} = {f}_t(c_i)^\top\, {f}_v(\mathcal{V}).
\end{equation}

Since all feature vectors are L2-normalized, these inner products equal cosine similarities. The vector $\mathbf{s}^{c} = (s_1^{c}, \ldots, s_K^{c})$ captures the intended structural importance of each shot relative to the global narrative (a purely text-derived reference), while $\mathbf{s}^{v} = (s_1^{v}, \ldots, s_K^{v})$ captures how well the generated video's visual content reflects each shot's intended contribution.

We quantify the structural alignment via the Spearman rank correlation coefficient between these vectors:
\begin{equation}
    M_\text{DSA}( \mathcal{V},\mathcal{C}) = \frac{\displaystyle\sum_{i=1}^{K}\bigl(R(s_i^{c}) - \bar{R}^{c}\bigr)\bigl(R(s_i^{v}) - \bar{R}^{v}\bigr)}{\sqrt{\displaystyle\sum_{i=1}^{K}\bigl(R(s_i^{c}) - \bar{R}^{c}\bigr)^2 \cdot \sum_{i=1}^{K}\bigl(R(s_i^{v}) - \bar{R}^{v}\bigr)^2}} 
\end{equation}
where $R(\cdot)$ denotes the rank function and $\bar{R}^{c}, \bar{R}^{v}$ are the respective mean ranks. The Spearman correlation is particularly appropriate here: it captures whether the relative ordering of shot importances is preserved in the generated video, rather than requiring exact magnitude agreement. This makes DSA inherently sensitive to disruptions in shot ordering and structural dynamics, precisely the long-range attributes that short-video metrics fail to capture.

\noindent\textbf{Long-Context Thinking with MLLM}.
While DSA captures temporal structural alignment, it cannot reason about fine-grained semantic attributes such as causal consistency across shots. To complement DSA, we design a multi-phase MLLM reasoning pipeline that employs deep thinking to assess long-range semantic quality.
As illustrated in Figure~\ref{fig:method}, our evaluation process unfolds in three key stages. (a) First, we employ an MLLM to segment the input video $\mathcal{V}$ into discrete shots and generate corresponding captions, denoted as $\hat{\mathcal{C}}$. (b) Second, we feed both the original $\mathcal{C}$ and the $\hat{\mathcal{C}}$ into an LLM for deep reasoning. This step systematically analyzes $\hat{\mathcal{C}}$ to detect potential structural errors (e.g., causal violations) and synthesizes a summary, $\mathcal{T}$. (c) Finally, we re-input all the aggregated information into the MLLM to compute the final metric, averaging the scores across $n$ independent sampling rounds:
\begin{equation}
    M_{\text{MLLM}} = \frac{1}{n}\sum_{i=1}^{n}\text{MLLM}(\mathcal{V}, \mathcal{C}, \mathcal{T}, \mathcal{B})
\end{equation}

where $\mathcal{B}$ denotes a reference set of human-annotated video cases utilized for in-context alignment.

The final Long-CODE metric integrates both dimensions:
\begin{equation}
    M_{\mathrm{Long\text{-}CODE}} = \alpha \cdot M_\text{DSA} + (1 - \alpha) \cdot M_{\mathrm{MLLM}},
\end{equation}
where DSA captures temporal-structural alignment from a shot-dynamics perspective, and $M_{\mathrm{MLLM}}$ provides deep semantic-level analysis of long-range narrative quality.

\begin{figure*}
    \centering
    \includegraphics[width=1\linewidth]{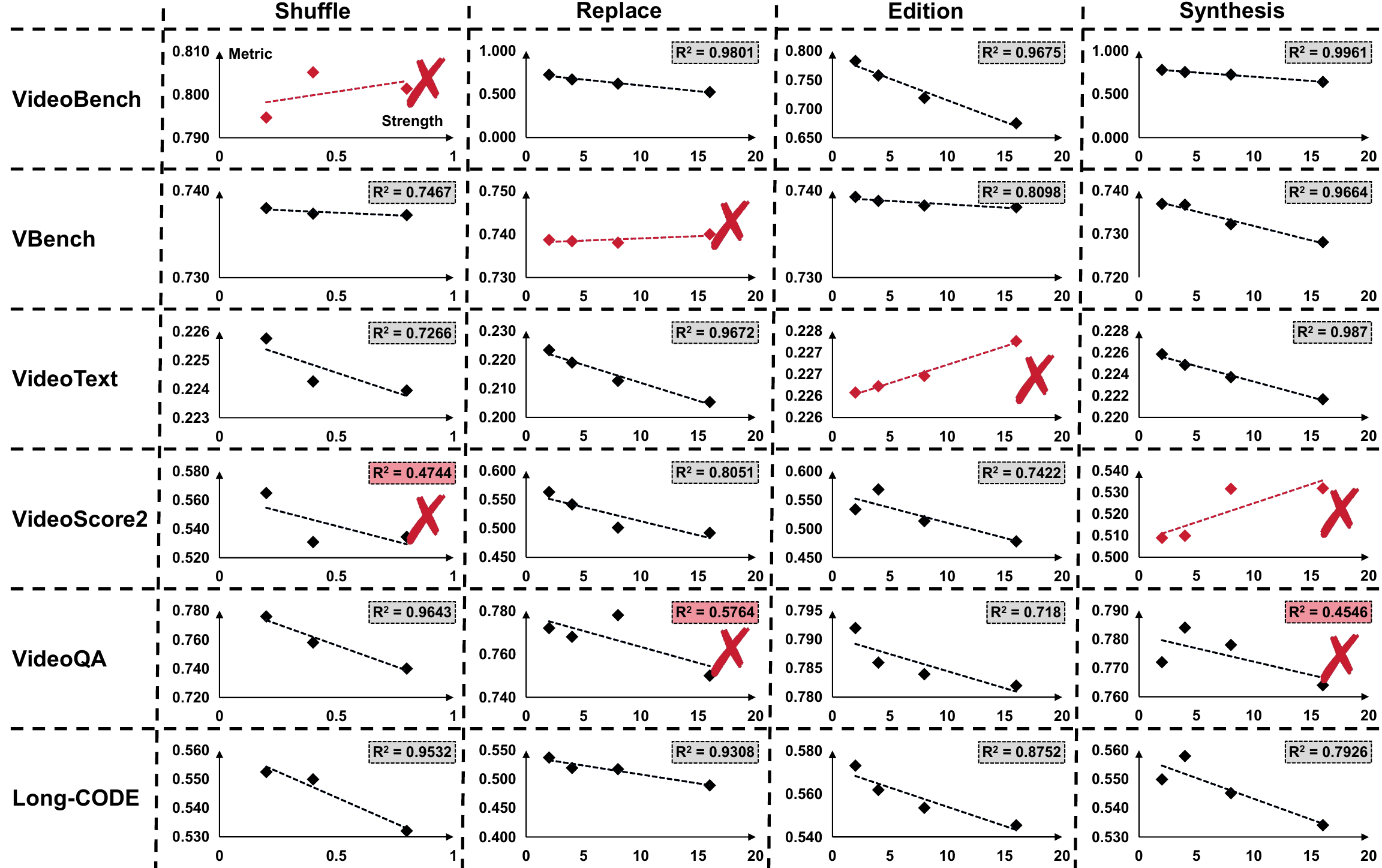}
    \caption{
    Correlation between metric scores and corruption strengths in the corruption tests. Since all evaluated benchmarks are higher-is-better metrics, their scores are expected to decrease as the corruption intensity increases. Consequently, if a metric exhibits a positive correlation with corruption strengths (indicated by the red lines in the figure, e.g., VideoBench in Shuffle) or yields a coefficient of determination $R^2 < 0.6$~\cite{draper1998applied}~(e.g., VideoQA in Replace), it is deemed insensitive to the given corruption (denoted by the red crosses).
    }
    \label{fig:corruption_test}
    \vspace{-12pt}
\end{figure*}
\subsection{Dataset}

\noindent\textbf{Long-CODE Benchmark Dataset.} To rigorously evaluate long-video generation metrics, we introduce a human-annotated dataset specifically targeting long-context attributes through a three-stage construction pipeline. First, an MLLM extracts a story seed from a single representative frame of a real-world video to generate a concise cinematic outline. Second, the MLLM decomposes this narrative into a sequence of structured shots, yielding the final prompt sequence $\mathcal{C} = (c_1, \ldots, c_K)$. Finally, multiple state-of-the-art models generate videos from $\mathcal{C}$, which human annotators evaluate strictly on long-range dimensions—such as causal consistency, temporal logic, and cross-shot identity persistence. By explicitly excluding per-shot visual quality from the evaluation rubric, this protocol isolates genuine long-context attributes from short-video fidelity confounds, establishing Long-CODE as a specialized complement to existing video generation assessment standards.






\section{Experiments}
\label{sec:exp}

\begin{table*}[htbp]
  \centering
    \renewcommand{\arraystretch}{1.1}
    \scalebox{1.1}{
\begin{tabularx}{0.9\textwidth}{l|>{\centering\arraybackslash}X>{\centering\arraybackslash}X>{\centering\arraybackslash}X>{\centering\arraybackslash}X>{\centering\arraybackslash}X>{\centering\arraybackslash}X|>{\centering\arraybackslash}X>{\centering\arraybackslash}X}
    \toprule
    \multicolumn{1}{c|}{\multirow{2}[2]{*}{Benchmark}} & \multirow{2}[2]{*}{VGoT} & \multirow{2}[2]{*}{HoloCine} & \multirow{2}[2]{*}{Wan2.2} & \multirow{2}[2]{*}{StoryMem} & \multirow{2}[2]{*}{Veo3.1} & \multirow{2}[2]{*}{Sora2} & \multicolumn{2}{c}{Overall} \\
          &       &       &       &       &       &       & Spearman & Pearson \\
    \midrule
    VBench & 0.013  & -0.035  & -0.069  & -0.090  & -0.171  & -0.105  & -0.023  & -0.104  \\
    \textcolor[rgb]{ .502,  .502,  .502}{\textit{\quad Motion}} & \textcolor[rgb]{ .502,  .502,  .502}{\textit{0.012 }} & \textcolor[rgb]{ .502,  .502,  .502}{\textit{-0.048 }} & \textcolor[rgb]{ .502,  .502,  .502}{\textit{0.124 }} & \textcolor[rgb]{ .502,  .502,  .502}{\textit{0.131 }} & \textcolor[rgb]{ .502,  .502,  .502}{\textit{0.074 }} & \textcolor[rgb]{ .502,  .502,  .502}{\textit{0.195 }} & \textcolor[rgb]{ .502,  .502,  .502}{\textit{0.137 }} & \textcolor[rgb]{ .502,  .502,  .502}{\textit{0.132 }} \\
    \textcolor[rgb]{ .502,  .502,  .502}{\textit{\quad Subject}} & \textcolor[rgb]{ .502,  .502,  .502}{\textit{0.228 }} & \textcolor[rgb]{ .502,  .502,  .502}{\textit{-0.006 }} & \textcolor[rgb]{ .502,  .502,  .502}{\textit{0.217 }} & \textcolor[rgb]{ .502,  .502,  .502}{\textit{0.193 }} & \textcolor[rgb]{ .502,  .502,  .502}{\textit{0.076 }} & \textcolor[rgb]{ .502,  .502,  .502}{\textit{0.165 }} & \textcolor[rgb]{ .502,  .502,  .502}{\textit{0.045 }} & \textcolor[rgb]{ .502,  .502,  .502}{\textit{0.047 }} \\
    \textcolor[rgb]{ .502,  .502,  .502}{\textit{\quad Dynamic}} & \textcolor[rgb]{ .502,  .502,  .502}{\textit{-0.152 }} & \textcolor[rgb]{ .502,  .502,  .502}{\textit{0.097 }} & \textcolor[rgb]{ .502,  .502,  .502}{\textit{-0.152 }} & \textcolor[rgb]{ .502,  .502,  .502}{\textit{-0.180 }} & \textcolor[rgb]{ .502,  .502,  .502}{\textit{-0.006 }} & \textcolor[rgb]{ .502,  .502,  .502}{\textit{-0.092 }} & \textcolor[rgb]{ .502,  .502,  .502}{\textit{0.014 }} & \textcolor[rgb]{ .502,  .502,  .502}{\textit{0.022 }} \\
    VideoScore2 & -0.023  & 0.061  & -0.140  & -0.069  & -0.160  & 0.004  & -0.010  & -0.028  \\
    VideoQA & 0.086  & -0.055  & -0.111  & 0.027  & -0.011  & \underline{0.127}  & -0.004  & -0.006  \\
    VideoText & \underline{0.138}  & -0.059  & \underline{0.204}  & 0.072  & 0.189  & 0.002  & 0.092  & 0.081  \\
    VideoBench & 0.069  & \underline{0.152}  & 0.042  &\underline{ 0.186 } & \underline{0.210}  & 0.108  & \underline{0.373}  &\underline{ 0.381}  \\
    \textcolor[rgb]{ .502,  .502,  .502}{\textit{\quad Object}} & \textcolor[rgb]{ .502,  .502,  .502}{\textit{0.081 }} & \textcolor[rgb]{ .502,  .502,  .502}{\textit{0.215 }} & \textcolor[rgb]{ .502,  .502,  .502}{\textit{0.042 }} & \textcolor[rgb]{ .502,  .502,  .502}{\textit{0.193 }} & \textcolor[rgb]{ .502,  .502,  .502}{\textit{0.304 }} & \textcolor[rgb]{ .502,  .502,  .502}{\textit{-0.016 }} & \textcolor[rgb]{ .502,  .502,  .502}{\textit{0.299 }} & \textcolor[rgb]{ .502,  .502,  .502}{\textit{0.294 }} \\
    \textcolor[rgb]{ .502,  .502,  .502}{\textit{\quad Scene}} & \textcolor[rgb]{ .502,  .502,  .502}{\textit{-0.010 }} & \textcolor[rgb]{ .502,  .502,  .502}{\textit{0.186 }} & \textcolor[rgb]{ .502,  .502,  .502}{\textit{0.026 }} & \textcolor[rgb]{ .502,  .502,  .502}{\textit{0.092 }} & \textcolor[rgb]{ .502,  .502,  .502}{\textit{0.131 }} & \textcolor[rgb]{ .502,  .502,  .502}{\textit{0.148 }} & \textcolor[rgb]{ .502,  .502,  .502}{\textit{0.300 }} & \textcolor[rgb]{ .502,  .502,  .502}{\textit{0.292 }} \\
    \textcolor[rgb]{ .502,  .502,  .502}{\textit{\quad Videotext}} & \textcolor[rgb]{ .502,  .502,  .502}{\textit{0.123 }} & \textcolor[rgb]{ .502,  .502,  .502}{\textit{0.121 }} & \textcolor[rgb]{ .502,  .502,  .502}{\textit{0.040 }} & \textcolor[rgb]{ .502,  .502,  .502}{\textit{0.133 }} & \textcolor[rgb]{ .502,  .502,  .502}{\textit{0.298 }} & \textcolor[rgb]{ .502,  .502,  .502}{\textit{0.141 }} & \textcolor[rgb]{ .502,  .502,  .502}{\textit{0.332 }} & \textcolor[rgb]{ .502,  .502,  .502}{\textit{0.327 }} \\
    Long-CODE & \textbf{0.708 } & \textbf{0.571 } & \textbf{0.835 } & \textbf{0.831 } & \textbf{0.796 } & \textbf{0.733 } & \textbf{0.765 } & \textbf{0.754 } \\
    \bottomrule
\end{tabularx}%
}
\vspace{4pt}
\caption{
Spearman~\cite{spearman} correlation coefficients between various benchmarks and human evaluations within each model.
``Overall'' denotes the Spearman and Pearson~\cite{pearson} correlation coefficients calculated by aggregating all samples. For the multi-dimensional benchmarks, VBench and VideoBench, we report the three metrics that exhibit the highest correlations. Specifically, in VBench~\cite{vbench}, ``Motion'', ``Subject'', and ``Dynamic'' denote the dimensions of motion smoothness, subject consistency, and dynamic degree, respectively. In VideoBench~\cite{videobench}, ``Object'', ``Scene'', and ``Videotext'' correspond to the dimensions of object class, scene, and video-text consistency, respectively. Bold numbers indicate the best performance, while underlined numbers denote the runner-up.
}
\vspace{-12pt}
\label{tab:human_corr}
\end{table*}%

\subsection{Models and Benchmarks}
\noindent \textbf{Long Video Generation Models.}
We evaluate 6 representative video generation models spanning both natively long-form generators and single-shot models extended via a sequential pipeline:

\begin{itemize}
    \item \textbf{VGoT}~\cite{vgot} is a training-free, modular framework.
    We generate each shot at 640p resolution with 25 denoising steps and a per-shot duration of 5\,s.
    \item \textbf{HoloCine}~\cite{holocine} holistically generates cinematic multi-shot narratives through Window Cross-Attention and Sparse Inter-Shot Self-Attention. Built upon the Wan2.2-T2V-A14B backbone, we produce 480$\times$832 videos at 15\,fps with 3 shots per generation chunk and 5\,s per shot.
    \item \textbf{Wan2.2}~\cite{wan} is Alibaba's open-source Mixture-of-Experts video diffusion model. We employ the TI2V-5B variant at 1280$\times$704 resolution with 25 sampling steps.
    \item \textbf{StoryMem}~\cite{storymem} adopts a Memory-to-Video design that maintains a dynamically updated keyframe memory bank.
    It fine-tunes Wan2.2 with lightweight LoRA (rank 128) and generates 832$\times$480 videos with 5\,s shots.
    \item \textbf{Veo3.1}~\cite{veo} is Google's commercial cinematic video engine. We produce 1920$\times$1080 clips of 8\,s each.
    \item \textbf{Sora2}~\cite{sora} is OpenAI's video generation model with synchronized audio and advanced physics simulation. We use the model configuration at 1280$\times$720, generating 15\,s clips.
\end{itemize}
Among these, VGoT, HoloCine, and StoryMem natively support multi-shot long-video generation.
For models that do not (i.e., Wan2.2, Veo3.1, and Sora2), we introduce a {long video generation pipeline}: each shot is generated conditioned on the current shot prompt and the last frame of the preceding shot, and all shots are then concatenated into a long video.
Each model generates 100 long videos from the Long-CODE Dataset, yielding a total of 600.

\noindent \textbf{Video Benchmarks.}
We compare Long-CODE against 5 established video evaluation methods that represent the predominant paradigms in the field.
\begin{itemize}
    \item \textbf{VBench}~\cite{vbench} is a comprehensive benchmark suite (CVPR 2024) that decomposes video quality into fine-grained dimensions, including subject consistency, background consistency, motion smoothness, dynamic degree, aesthetic quality, and imaging quality; we evaluate all applicable dimensions for long videos.
    \item \textbf{VideoScore2}~\cite{videoscore2} is a fine-tuned MLLM evaluator (based on Qwen2.5-VL-7B) that explicitly scores three dimensions (i.e., visual quality, text-to-video alignment, and physics/common-sense consistency) with chain-of-thought reasoning.
    \item \textbf{VideoQA}~\cite{song2026vqqa,lin2024evaluating} assesses narrative comprehension through multiple-choice questions: we auto-generate 5 questions per sample from the storyline and use Qwen3-VL-32B-Instruct~\cite{qwen3vl} to answer them over the generated video, reporting accuracy.
    \item \textbf{VideoText}~\cite{liu2024evalcrafter,videoclip} computes the per-shot cosine similarity between VideoCLIP-XL~\cite{videoclip} video features and text features and averages across shots.
    \item \textbf{VideoBench}~\cite{videobench} (CVPR 2025) employs a Qwen3-VL-32B-Instruct judge to score videos across 9 quality dimensions with few-shot, chain-of-query prompting.
\end{itemize}
\begin{figure*}
    \centering
    \includegraphics[width=1\linewidth]{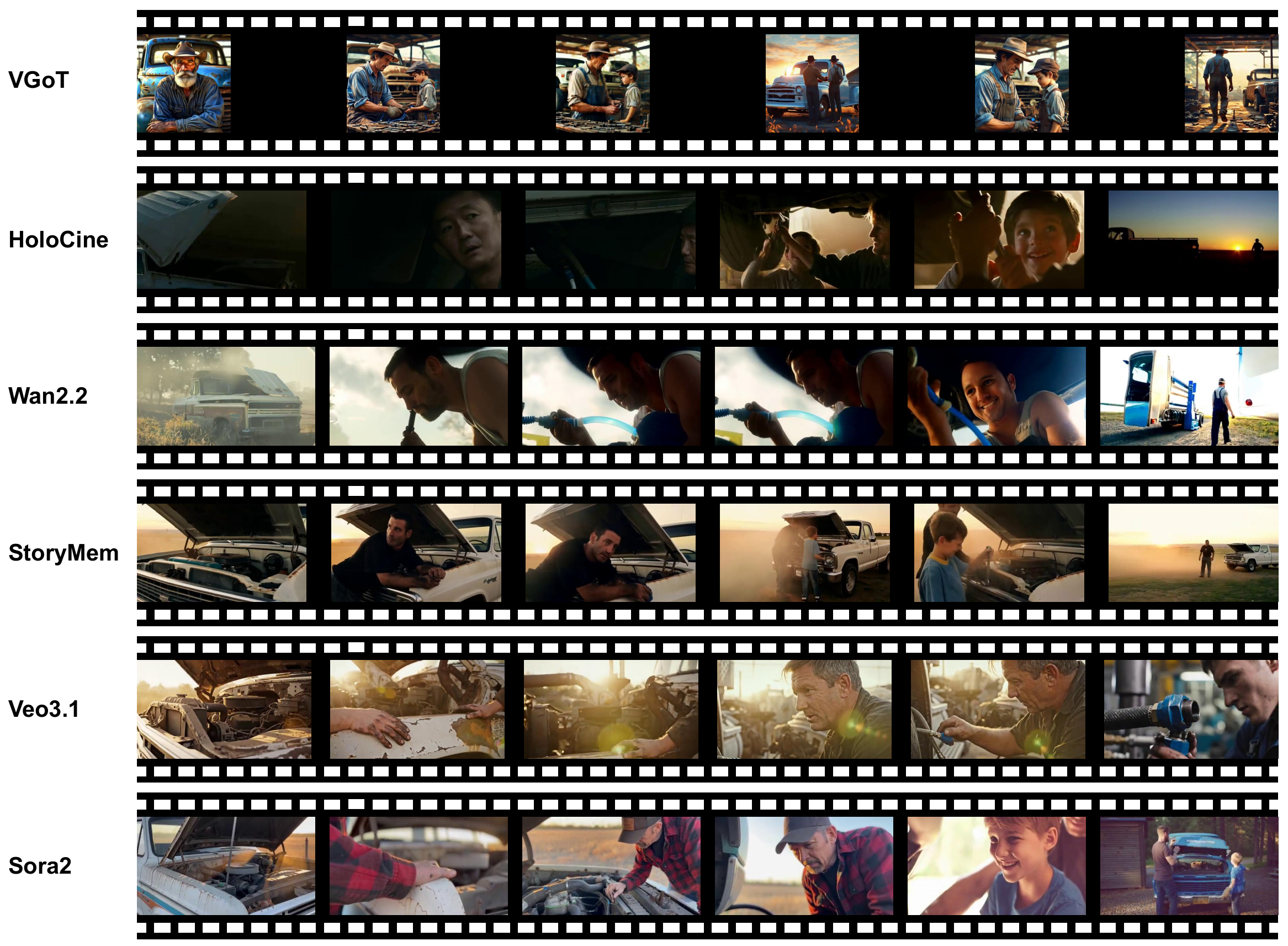}
    \caption{A case study of different long video generation models on the Long-CODE dataset. Storyline is ``At dawn, a mechanic works on the dusty engine of an old white pickup truck. While checking a hose, he remembers the day his late son helped him install it. With a sigh, he pours in fresh coolant, keeping the old truck running in honor of his son's memory.''}
    \label{fig:case}
    \vspace{-12pt}
\end{figure*}

\subsection{Implementation}
\label{subsec:implementation}
\noindent \textbf{Corruption Tests.} 
To expose the limitations of current short-video metrics, we devise four long-range attribute corruption tests and apply them to 100 real videos from FineVideo~\cite{finevideo}.

\begin{itemize}
    \item \textbf{Shuffle:} Segments the video into 10-second blocks and randomly permutes them, with the proportion of shuffled blocks varying across \{0.2, 0.4, 0.8\}.
    \item \textbf{Replace:} Substitutes chosen blocks with the most cosine-similar 10-second segments retrieved from a 100-video candidate pool, encoded via VideoCLIP-XL~\cite{videoclip}. The number of replaced blocks varies across \{2, 4, 8, 16\}.
    \item \textbf{Edition:} Utilizes VACE~\cite{jiang2025vace} (1.3B) for depth-guided editing to modify character identities (e.g., gender swapping) while maintaining the original scene layout. We randomly select \{2, 4, 8, 16\} blocks (81 frames each) for editing.
    \item \textbf{Synthesis:} Generates captions for randomly selected blocks using Qwen3-VL-8B-Instruct, rewrites them into surreal descriptions, and resynthesizes the blocks using Wan2.2 TI2V-5B ($1280\times704$ resolution, 25 denoising steps). The number of resynthesized blocks varies across \{2, 4, 8, 16\}.
\end{itemize}

\noindent \textbf{Long-CODE Metric.}
The Long-CODE metric sets $\alpha = 0.5$ balances the two complementary sub-metrics.
${M}_{\text{DSA}}$ uses VideoCLIP-XL~\cite{videoclip}
to extract features, and finally normalized to $[0,1]$.
${M}_{\text{MLLM}}$ (\emph{MLLM-as-Judge}) leverages Qwen3-VL-32B-Instruct~\cite{qwen3vl} describes each shot from the sampled frames; analyzes quality from four perspectives (i.e., visual richness, temporal transition, object/background consistency, and logic errors), producing a structured summary; the target video is compared against 3 reference videos (randomly drawn from a bank of 15 human-scored samples) and scored on a 1--5 scale.
This process is repeated for 3 rounds with increasing temperature ($t\in \{0.3, 0.4, 0.5\}$) and different reference selections; the final score is the average across rounds, normalized to $[0, 1]$.

\noindent \textbf{Long-CODE Dataset.}
We source 100 diverse real-world videos from the FineVideo~\cite{finevideo} dataset as seed material.
For each video, we extract a random frame and prompt Qwen3-VL-32B-Instruct~\cite{qwen3vl} to produce a cinematic story outline.
The outline is then expanded into a sequence of shot descriptions with duration and cut-type annotations, targeting a total duration of 60--120\,s.
To ensure temporal regularity, each shot is normalized to a fixed duration of 5\,s.
This yields a structured prompt comprising 12--24 shots per sample, each annotated with a textual description, duration, and cut indicator.

Human evaluation is conducted on three long-range dimensions that are orthogonal to short-video visual quality: {narrative accuracy} (whether the generated video faithfully renders the intended storyline), {long-range causality} (whether cause and effect relationships hold across distant shots), and {cross-scene identity consistency} (whether characters and key objects maintain coherent appearance throughout the video).
Each of the 600 generated videos (100 samples $\times$ 6 models) is rated by 8 human annotators on a 1--5 Likert scale for these three dimensions.

\subsection{Result}
\noindent\textbf{Q1.} \textbf{\textit{Are long-context attributes orthogonal to short-video dimensions?} }

\noindent\textbf{A1.} Long-context attributes are orthogonal to short-video dimensions. As shown in Figure~\ref{fig:corruption_test}, we systematically apply long-range structural corruptions to long videos and measure the sensitivity of existing metrics. A reliable metric should exhibit a monotonic decrease as corruption intensity increases. However, established short-video metrics fail this fundamental test: VideoBench shows a {positive} correlation with corruption strength under the Shuffle test. 
Similarly, VBench, VideoText, and VideoScore2 failed the Replace, Edition, and Synthesis tests, respectively. Furthermore, although VideoQA demonstrates a negative correlation across all corruption tests, its $R^2 < 0.6$ in the Replace and Synthesis tests, reflecting a negligible sensitivity to content-level perturbations~\cite{draper1998applied}.

Qualitative analysis further corroborates this orthogonality. As shown in the Figure~\ref{fig:case}, each individual shot can exhibit high visual quality, yet the overall long video suffers from severe cross-scene inconsistencies, including character identity drift, broken causal chains, and incoherent narrative progression.

As shown in Table~\ref{tab:human_corr}, quantitative analysis at scale confirms the above findings. When evaluated against human judgments specifically annotated for long-range attributes, short-video metrics exhibit near-zero or even negative correlations: VBench achieves an overall Spearman correlation of merely $-0.023$, VideoScore2 yields $-0.010$, and VideoQA produces $-0.004$. Even the best sub-dimensions of VBench, Motion ($0.137$), Subject ($0.045$), and Dynamic ($0.014$), remain negligible. These results confirm that metrics designed for short-video quality, focusing on local temporal coherence, are inherently blind to long-range structural disruptions.

\begin{table}[t]
  \centering
    \renewcommand{\arraystretch}{1.1}
    \scalebox{1.1}{
\begin{tabularx}{0.9\linewidth}{l|>{\centering\arraybackslash}X>{\centering\arraybackslash}X>{\centering\arraybackslash}X>{\centering\arraybackslash}X}
    \toprule
          & Narrative & Causality & Consistency \\
    \midrule
    $M_{\text{DSA}}$   & 0.259  & 0.214  & 0.219  \\
    $M_{\text{MLLM}}$  & 0.637  & 0.592  & 0.604  \\
    \midrule
    $M_{\text{Long-CODE}}$ & \textbf{0.685 } & \textbf{0.620 } & \textbf{0.626 } \\
    \bottomrule
\end{tabularx}%
}
\caption{
Ablation studies of the proposed Long-CODE metric across different human evaluation dimensions. The human evaluation dimensions ``Narrative'', ``Causality'', and ``Consistency'' denote narrative accuracy, long-range causality, and cross-scene identity consistency, respectively (Section~\ref{subsec:implementation}).
}
 \vspace{-24pt}
\label{tab:ablation_human}
\end{table}%

\noindent\textbf{Q2.} \textbf{\textit{Does Long-CODE align with human preferences in evaluating long videos? } }

\noindent\textbf{A2.} Long-CODE align with human preferences in evaluating long videos. As reported in Table~\ref{tab:human_corr}, Long-CODE obtains the highest Spearman correlation with human judgments across all six evaluated models without exception: VGoT ($0.708$), HoloCine ($0.571$), and etc.. These correlations are consistently strong ($>0.57$) across models spanning diverse architectures and generation paradigms, demonstrating the robustness of our metric.

The overall Spearman and Pearson correlations of Long-CODE are $\mathbf{0.765}$ and $\mathbf{0.754}$, respectively. The strongest baseline, VideoBench, achieves only $0.373$ (Spearman) and $0.381$ (Pearson), less than half of Long-CODE's correlation. All other baselines fall below $0.1$ in overall correlation. This constitutes a relative improvement of over $105\%$ in Spearman correlation over the best existing metric.
While certain baselines exhibit moderate per-model correlations (e.g., VideoText achieves $0.204$ on Wan2.2), they fail to maintain consistent performance across models. In contrast, Long-CODE's {minimum} per-model correlation ($0.571$ on HoloCine) already exceeds the {maximum} per-model correlation of any baseline. This underscores the superior generalizability of our metric.

\noindent\textbf{Q3.} \textbf{\textit{What specific qualities of long videos do the two metrics within Long-CODE respectively evaluate? } }

\noindent\textbf{A3.} The two component metrics within Long-CODE, Dynamic Structure Alignment $M_{\text{DSA}}$ and $M_{\text{MLLM}}$, capture complementary dimensions of long-video quality, and their combination yields consistent improvements.
As shown in Table~\ref{tab:ablation_human}, $M_{\text{MLLM}}$ dominates on all three semantic evaluation dimensions, achieving correlations of $0.637$ (Narrative), $0.592$ (Causality), and $0.604$ (Consistency). This is expected, as the MLLM component explicitly reasons about narrative content, causal relationships, and cross-scene identity. In contrast, $M_{\text{DSA}}$ alone yields lower correlations ($0.259$, $0.214$, $0.219$), since it operates on structural shot dynamics rather than semantic content. Crucially, the combined metric $M_{\text{Long-CODE}}$ consistently surpasses both individual components: $0.685$ ($+0.048$ over $M_{\text{MLLM}}$ on Narrative), $0.620$ ($+0.028$ on Causality), and $0.626$ ($+0.022$ on Consistency). 
This complementary sensitivity profile confirms the design motivation: $M_{\text{DSA}}$ evaluates temporal-structural quality, while $M_{\text{MLLM}}$ evaluates semantic-content quality.
Their fusion in $M_{\text{Long-CODE}}$ yields a comprehensive metric sensitive to both corruption categories.


\noindent\textbf{Q4.} \textbf{\textit{Why is Dynamic Structure Alignment~(DSA) effective in evaluating long-video dynamics?} }

\noindent\textbf{A4.} From the formulation perspective.
The key insight behind DSA is a decoupling of {what each shot should contribute} from {what it actually contributes} in the generated video. Recall that $\mathbf{s}^{c}$ encodes the intended structural importance of each shot relative to the global narrative purely from text, while $\mathbf{s}^{v}$ reflects how the generated video's visual content realizes each shot's intended contribution. By comparing these two profiles via Spearman rank correlation, DSA evaluates whether the {relative ordering} of shot importance is faithfully preserved, rather than demanding exact magnitude agreement. 
DSA is {inherently sensitive to long-range structural disruptions}: any perturbation that alters the temporal arrangement of shots (e.g., narrative shuffling) directly rearranges $\mathbf{s}^{v}$ while leaving $\mathbf{s}^{c}$ unchanged, producing a measurable drop in $M_{\text{DSA}}$. 

From the empirical perspective. 
The ablation in Table~\ref{tab:ablation_human} reveals that even though $M_{\text{DSA}}$ alone achieves moderate human correlation, it consistently \textit{improves} the combined metric: $M_{\text{Long-CODE}}$ outperforms $M_{\text{MLLM}}$ by $+3.3\%$--$7.5\%$ across all three human evaluation dimensions, confirming that DSA captures structural information complementary to semantic evaluation.

\section{Conclusion}
In this paper, we identified a critical gap in the evaluation of Video Generation Models (VGMs) as they scale to generate longer content. We theoretically and empirically demonstrated that short-term visual perception and long-context attributes are fundamentally orthogonal dimensions. To address the insensitivity of traditional short-video metrics to structural inconsistencies, we introduced a suite of long-video attribute corruption tests. Furthermore, we developed a novel evaluation metric based on shot dynamics and the deep reasoning capabilities of Multimodal Large Language Models. Alongside this, we introduced Long-CODE, a specialized dataset designed to benchmark long-video evaluation, with human annotations isolated specifically to genuine long-range characteristics. Extensive experiments show that our proposed metrics achieve state-of-the-art correlation with human judgments. Ultimately, our metric and benchmark seamlessly complement existing short-video standards, establishing a holistic and unbiased evaluation paradigm for video generation models.

\bibliographystyle{ACM-Reference-Format}
\bibliography{main}

\end{document}